# Audio Simulation for Sound Source Localization in Virtual Environment

Yuan Yi Di[†]
*Home Team Science and Technology Agency*
Singapore
Yuan_Yidi@htx.gov.sg

Wong Swee Liang[†]
*Home Team Science and Technology Agency*
Singapore
Wong_Swee_Liang@htx.gov.sg

Jonathan Pan*
*Home Team Science and Technology Agency*
Singapore
Jonathan_Pan@htx.gov.sg

*Abstract*—Non-line-of-sight localization in signal-deprived environments is a challenging yet pertinent problem. Acoustic methods in such predominantly indoor scenarios encounter difficulty due to the reverberant nature. In this study, we aim to locate sound sources to specific locations within a virtual environment by leveraging physically grounded sound propagation simulations and machine learning methods. This process attempts to overcome the issue of data insufficiency to localize sound sources to their location of occurrence especially in post-event localization. We achieve 0.786 ±0.0136 F1-score using an audio transformer spectrogram approach.

## I. Introduction

Localization, which is to determine the position of an individual or an object within a specific space, is a crucial task in the line of work for public safety in situations where a threat/rescue target is required to be tracked and located during emergencies. In outdoor environments, localization tasks predominantly leverage the Global Navigation Satellite Systems (GNSS) such as the Global Positioning System (GPS) and the likes to provide meter-accurate positioning information over large areas using trilateration of the time-of-flight of signals transmitted from satellites to the receiver [1]. However, the system loses accuracy and reliability in indoor environments due to signal attenuation or shielding by building materials, complex indoor environments, and signal interferences due to multipath propagation, etc. [2].

To enable more accurate indoor localisation, research and rapid adoption of Internet of Things (IoT) networks consisting of high-penetration wireless technologies, such as ultra-wideband (UWB) [3], WiFi [4], FM radio [5], Ultrasound [6], Bluetooth [7], LoRA [8], etc., mobile devices, and sensors have enabled fairly accurate localisation of these devices and their carrier within the network [9], [10]. This is achieved through the transmission and receiving of relative position data among predetermined fixed nodes or mobile computing devices (such as smartphones, watches, beacons, vehicles, drones, etc.). These methods assume pre-installation of required devices or presence of the device on the target to be located. However, in a wireless signal-deprived environment or in the absence of these devices both in the environment and on the target, makes localization, especially non-line-of-sight localization, an even harder problem to solve.

Interestingly, in nature, sound is an abundant signal that is widely used as a means of localization. For example, echolocation is a commonly used technique by certain animals and even humans to identify their environments e.g. distance, obstacles, and space [11]. Research in sound source localization (SSL), or acoustic sound localization (ASL), is a long-standing effort that aims to achieve similar functions as echolocation, leveraging sound to estimate the positions of sound sources based on audio data collected by receivers e.g. microphones [12]. Classically, audio data is analyzed using signal processing techniques that require an array of microphones where the geometry of the array is known for the estimation of direction of arrival (DoA) through estimating the time-difference of arrival (TDoA) between the microphones [12], [13]. However, this method only estimates the azimuth and elevation angles, but not the source-receiver distance [14]. And despite a significant amount of research in SSL, the localization performance is still often unsatisfactory when complex factors such as noise, reverberation, and interference from sound mixtures are involved [15].

In the past decade, research into data-driven deep learning has shown better localization accuracy compared to classical signal processing methods [16], [17], [18]. However, very often these audio data are collected in carefully controlled environments which results in relatively small datasets that are environment-specific and difficult for downstream tasks such as machine learning model training to generalize to other environments. In time-constraint scenarios, the resources and time available to collect this data is also limited. One possible method to expand the dataset is through simulation-to-reality (sim-to-real) transfer of sound data using sound propagation modelling (SPM), which is to emulate realistic, high-fidelity acoustics that align with the specified environment conditions, without needing real-life data collection. This is especially useful for localization in post-event analysis through audio recordings. In recent years, the development of SPM has demonstrated hyper-realistic interactive simulation in virtual spaces [19], [20]. Chen et al. have also shown that SoundSpace2.0 simulation engine is capable of realistic audio generation in a given room three-dimensional (3D) mesh matching that to real acoustic measurements of the same physical room [20].

Therefore, we attempt to develop a workflow that allows for localization of non-line-of-sight sound sources in signal-scarce environments, especially for post-event analysis of the data, through a sim-to-real method. This includes highly realistic generation of reverberation of sound in a virtual 3D environment, rapid generation of 3D environments, and training of models for accurate positioning of the generated sound. As the first step, in this study, we explore the possibility of using simulated sounds to train machine learning models to localize sound sources within a single virtual indoor setting.

## II. Related Work

Methods such as optical-acoustic observation [21] and diffraction and reflection directions [22] have been reported for simple non-line-of-sight (NLOS) indoor sound source positioning in single room and simple scenarios [22], [23]. Multi-room NLOS localization is more challenging due to pollution by high noise-to-signal ratio, reverberation etc. and commonly used acoustic features such as DOA and TDOA give inadequate information. The use of deep neural networks (DNNs) has proved to be effective in extracting useful features from noisy audio data [25], [26]. However, the lack of training data has made it difficult for useful applications. Liu et. al. [27] proposed the use of self-populated data from active robot exploration in complex multi-room environment. The robot can collect data through self-supervised exploration of the rooms and update the learned model on-the-fly. The acoustic features obtained using autoencoder and geometric information were used to train the DNN model. The model was able to reach 90% accuracy with 120 samples. However, this method will require physical collection of data.

## III. METHOD

Our goal in this study is to perform localization of sound sources within a predetermined house using simulated audio. Here, we leverage the SoundSpace2.0 [20] framework and Habitat-Sim and Habitat-Lab simulation engine [25] to construct a pipeline for simulating the received sound in a virtual 3D room. With the convenience of simulating all possible source and receiver positions, sufficient data is generated to effectively train a machine learning model. Thus, this enables us to use audio spectrograms, instead of the more commonly used DoA, as the training data.

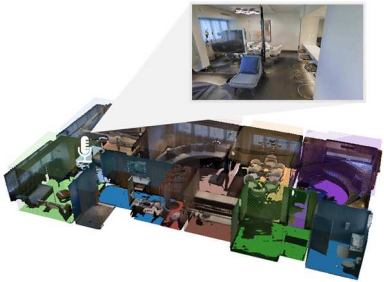

Fig. 1. MP3D room (code: 17DRP5sb8) and position of sound receiver. Insert: view of the agent shows the direction in which the sound receiver faces.

### A. Rendering Pipeline

SoundSpace2.0 integrates the audio propagation engine (RLR-Audio-Propagation) into Habitat-Sim. Fig. 1 illustrates the simulation pipeline. SoundSpace2.0 computes the room impulse response (RIR) by taking in the scene mesh data that is processed by Habitat, and user-specified source and receiver locations in the 3D environment. This is achieved using the geometric method of Monte Carlo bidirectional path-tracing algorithm [19]. Unlike the computationally intensive wave function based RIR generation, geometric path tracing is favored for its simplicity and ability to handle complex environments and generate unbiased results. Traditional path tracing is mono-directional with paths emitting from the sound source towards the receiver. Bidirectional path tracing has subpaths emitting from the emitter and receiver ends and builds connections between the subpaths to construct the final paths.

The dry sound, the original sound that has not interacted with the 3D environment, was normalized using Audacity. The peak amplitude of the dry sound file is normalized to -1.0 dB and the waveform centered on 0.0 vertically. The perceived loudness of the file was then normalized to -15.0 LUFS. The normalized dry sound was exported as a 16 bit PCM WAV file. The generated RIR was then convolved with the normalized dry sample sound to produce the final reverbed sound, from which spectrogram with time-frequency information was then generated using Short-time Fourier Transform. The spectrograms were used as input data to train the machine learning models.

### B. Dataset

A single-level model house which is well-integrated with the Habitat environment was obtained from the Matterport3D (MP3D) dataset [26]. The model house is 15.6 m x 4.4 m x 8.3 m in dimensions. The dry audio track was obtained from SoundSpace2.0 repository. For room-based localization, 640 spectrograms of reverb-convolved audio were generated using our described rendering pipeline. The indoor environment consists of 10 different rooms and the microphone was situated only in one of the rooms with a minimal line of sight to the sound sources. For coordinate-based localization, 512 spectrograms of reverb-convolved audio were generated instead.

### C. Machine Learning Models

Instead of the commonly used DoA method, audio spectrograms were directly used as training data. Convolutional Neural Networks (CNN) is a commonly used method for learning spectrogram data. Hence, we first attempted localization using CNN models trained on the audio spectrograms for either classification or regression outputs. The models were constructed using keras. The CNN model comprises 2 convolutional blocks (2D convolutional layer, pooling layer and batch normalization) followed by 3 densely connected layers, with the final output layer having a SoftMax activation function for classification or a vector output for regression. The output of the classification model were probability scores with the class having the highest score indicating the room where the sound originated from. The output of the regression model was a vector comprising the x, y coordinates of the sound source.

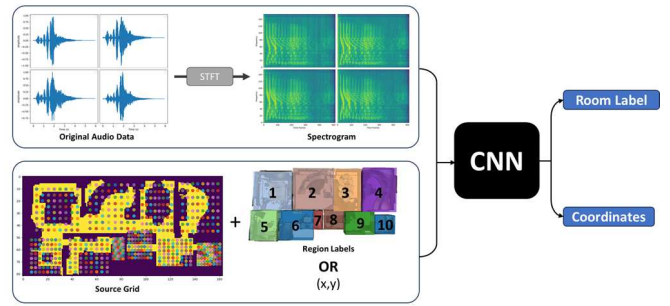

Fig. 2. Schematic overview of the data for predicting the specific room and coordinates, respectively, and in which the sound source resides.

We compared the CNN classification model with a different model modified from an existing Audio Spectrogram Transformers (AST) by Gong et al. [27] that was pre-trained on Audio Set [28]. The final dimension of the output layer of the original AST model was replaced from the original 527 classes to 10 classes, representing the number of rooms. We then finetuned the pre-trained AST on the simulated data. The architecture of the model is shown in Figure 3.

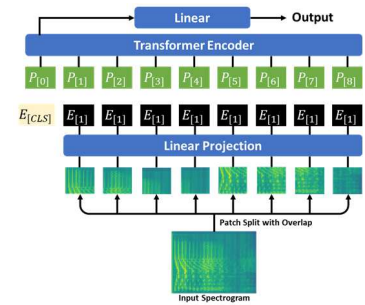

Fig. 3. Schematic overview of the AST for predicting the specific room in which the sound source resides.

The audio spectrograms were treated as 2D images, with the horizontal spatial coordinates being the time frames of the audio and the vertical spatial coordinates as the audio frequencies. Pixel intensities are the associated amplitudes. 5-fold cross-validation was done, in which 80% of the total data is trained on while the remaining 20% is used for evaluation to account for data variation.

### D. Evaluation Metrics

For the classification model, the trained model was evaluated for their performance based on recall, precision and F1-scores. The following equations describes these metrics:

$$Precision = \frac{TP}{TP+FP} \quad (1)$$

$$Recall = \frac{TP}{TP+FN} \quad (2)$$

$$F1\ score = 2 \times \frac{Precision \times Recall}{Precision+Recall} \quad (3)$$

where the variables are defined as:

True positive (TP): number of correct model predictions on the positive class, False positive (FP): number of incorrect model predictions on the positive class, False negative (FN): number of incorrect model predictions on the negative class.

For the regression model, model performance was evaluated based on mean squared error values.

## IV. EXPERIMENTS AND DISCUSSION

### A. Localization by coordinates with single receiver.

Ideally, the geographical coordinates provide a direct indication of the location of the sound source and allow for easy generalization to various locations irrespective of their complexity. Hence, we first attempted to predict the (x,y) coordinates of the sound source.

Due to the dimension of the house, the rooms were densely sampled with a distance of 0.52 m in between point sources across the entire room at a constant Z-plane of 1.7 m to offer sufficient point cloud. A test set of 100 source points, large enough to cover all rooms, was also generated using similar method but does not coincide with the training set. The Euclidean distances between the predicted (x,y) coordinates of the test set and the actual coordinates are then calculated to check the prediction accuracy. In Fig. 4, the radius represents the leniency in which we consider the prediction to be accurate within a certain distance away from the actual coordinate. It is found that, the minimum leniency, which we set at 50%, is at 3.4 m.

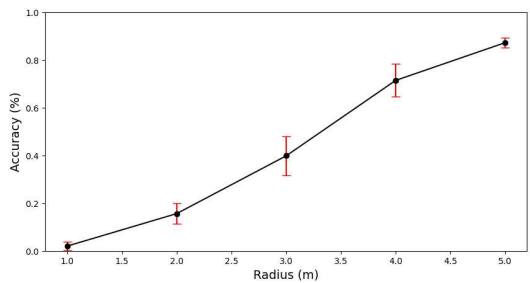

Fig. 4. Accuracy of model prediction for the coordinates of sound source.

### B. Localization of source by spatial regions with single receiver.

To improve the prediction accuracy, we reformat the prediction into a classification problem, where the location will be indicated as a spatial region instead of a single geometric point. The room model was separated into 10 regions and each region was densely sampled with an equal number of point sources laid in uniform grids of 8 x 8. This is to avoid the problem of unbalanced sampling during model training. The boundaries of the regions were scaled to minimize ambiguity at the boundaries of the spaces. A total of 640 source points was laid out within the grid and will be referred to as the source set.

We first trained the 5-fold CNN model with a complete source set convolved with a single type of dry sound and the corresponding room name label. The model architecture is shown in Fig. 2. Categorical cross-entropy loss function was selected for multi-class classification and categorical accuracy as the metrics to evaluate the effectiveness of the model. The same method wass employed for finetuning the AST model.

A test set of 250 source points that do not coincide with the source set was simulated similarly in the Habitat environment, while retaining the same receiver position and convolved with the same dry sound. Their room location was predicted using the CNN model and the results provided in Table 1. The room-based localization evidently performed better compared to coordinate-based prediction. This is perhaps due to greater leniency given in terms of area size as well as characteristic reverb within the rooms. Comparing the CNN model with the AST based model reveals that the AST model performs much better for either receiver locations. This shows that the model performance is robust against receiver location. This could be due to the pre-training that the AST model went through and the model ability to understand the relative positions of the spectrogram, instead of the translational equivariance of the CNN model.

TABLE I. CLASSIFICATION RESULTS.

| Model | F1 Score | Precision | Recall |
|---|---|---|---|
| AST | 0.786 ± 0.014 | 0.812 ± 0.013 | 0.784 ± 0.015 |
| CNN | 0.594 ± 0.019 | 0.626 ± 0.031 | 0.656 ± 0.027 |

## V. CONCLUSION AND FUTURE DIRECTIONS

In this study, we explored the possibility of using simulated RIR to locate sound source in a virtual environment. We found that directly locating the sound source via geographical coordinates remains a challenging task. However, localizing to a specific spatial region proves to be easier and more accurate. As discussed by Chen et al. [20], there are still some limitations of the SoundSpace2.0 simulation engine that make complete realism challenging, this is a valuable platform to generalize sound propagation to different environments beyond the physical limitations. And this current study is our first step towards sim-to-real adaptation for sound localization. There are a few of the future works we foresee can build up or improve the current state of work:

### A. Adaptation to dynamic scenarios.

The current study looks at static sources. However, as the sound source may move in real scenarios, localizing and tracking of moving sound source is critical for effective localization in real-life scenarios [15]. Furthermore, we only used single level indoor environment for the simulation. The numbers of levels may affect the simulation and data-processing in different ways, which is crucial for generalizing the model.

*B. Sound separation for mixtures.*

The current audio dataset consists of only pure sounds. Hence, the model performed poorly in the instances of sound mixtures. In future development, source splitting is necessary to deal with sounds in real-life data [14].

*C. Dereverberation of actual audio.*

In this study, we simulated the reverbs of audios and convolved the reverb with stock dry sounds. However, in real life, the production of sounds is spontaneous, and the collection of dry audio is not possible during actual applications. Hence, dereverberation is necessary to directly obtain the reverbs for localization and the dry audio for training.

*D. Building of 3D virtual environment based on real-world buildings blueprints.*

In order for real-to-sim transfer of the machine learning models trained in simulation, it is imperative that an accurate 3D model of the real-life indoor environment can be constructed.

*Data Availability.* The data that support the figures within this paper and other finding of this study are available from the authors upon reasonable request. Requests for materials should be addressed to Y. Yuan and S.L. Wong.